\documentclass{article}
\usepackage{spconf}
\ninept

\usepackage[utf8]{inputenc} 
\usepackage[T1]{fontenc}    
\usepackage{hyperref}       
\usepackage{url}            
\usepackage{booktabs}       
\usepackage{amsfonts}       
\usepackage{nicefrac}       
\usepackage{microtype}      
\usepackage{siunitx}
\usepackage{enumerate}
\usepackage{float}
\usepackage{caption}

\usepackage{lipsum}

\hypersetup{allcolors=blue,colorlinks=true}

\usepackage{graphicx}
\usepackage{amsmath}
\usepackage{amsthm}
\usepackage{amssymb}
\usepackage{amsfonts}
\usepackage{mathrsfs}
\usepackage{multirow}
\usepackage{algorithm}
\usepackage{subfigure}
\usepackage{thm-restate}
\usepackage{mathtools}
\usepackage{comment}
\usepackage{multicol}
\usepackage{bm}
\newcommand{\R}{\mathbb{R}}
\newcommand{\B}{\mathbb{B}}
\newcommand{\bA}{\bm{A}}
\newcommand{\bB}{\bm{B}}
\newcommand{\bDelta}{\bm{\Delta}}
\newcommand{\bD}{\bm{D}}
\newcommand{\bE}{\bm{E}}
\newcommand{\bI}{\bm{I}}
\newcommand{\bS}{\bm{S}}

\newcommand{\bU}{\bm{U}}
\newcommand{\bV}{\bm{V}}
\newcommand{\bW}{\bm{W}}
\newcommand{\bX}{\bm{X}}
\newcommand{\bY}{\bm{Y}}
\newcommand{\bOmega}{\bm{\Omega}}
\newcommand{\bu}{\bm{u}}
\newcommand{\bv}{\bm{v}}

\newcommand{\bZ}{\bm{Z}}

\newcommand{\V}{\mathcal{V}}

\usepackage{bbm}

\newcommand{\bzero}{\mathbf{0}}

\DeclareMathOperator*{\argmin}{arg\,min}
\DeclareMathOperator*{\argmax}{arg\,max}
\DeclareMathOperator{\prox}{\textsf{prox}}

\DeclareMathOperator{\Retr}{\textsf{Retr}}
\DeclareMathOperator{\SVD}{\textsf{SVD}}
\DeclareMathOperator{\Tr}{Tr}

\usepackage{pifont}
\newcommand{\cmark}{\ding{51}}%
\newcommand{\xmark}{\ding{55}}%

\newtheorem{theorem}{Theorem}

\newtheorem{lemma}{Lemma}

\newcommand{\pp}{\phantom{+}}

\newcommand{\bSu}{\bS_{\bu}}
\newcommand{\bSv}{\bS_{\bv}}

\usepackage[natbib=true,
            bibstyle=ieee,
            citestyle=numeric-comp,
            uniquename=false,
            uniquelist=false,
            maxcitenames=2,
            mincitenames=1,
            mincrossrefs=99]{biblatex}
\renewbibmacro{in:}{\ifentrytype{article}{}{\printtext{\bibstring{in}\intitlepunct}}}
\DefineBibliographyStrings{english}{url = Available at}

\title{Multi-Rank Sparse and Functional PCA \\ Manifold Optimization and Iterative Deflation Techniques}

\name{Michael Weylandt\thanks{MW acknowledges support from the NSF Graduate Research Fellowship Program under grant number 1450681.}}
\address{Department of Statistics\\
         Rice University\\
         Houston, TX 77005\\
         \href{mailto:michael.weylandt@rice.edu}{michael.weylandt@rice.edu}}

\begin{document}
\begin{refsection}
\maketitle

\begin{abstract}
We consider the problem of estimating multiple principal components using the recently-proposed Sparse and Functional Principal Components Analysis (SFPCA) estimator. We first propose an extension of SFPCA which estimates several principal components simultaneously using manifold optimization techniques to enforce orthogonality constraints. While effective, this approach is computationally burdensome so we also consider iterative deflation approaches which take advantage of existing fast algorithms for rank-one SFPCA. We show that alternative deflation schemes can more efficiently extract signal from the data, in turn improving estimation of subsequent components. Finally, we compare the performance of our manifold optimization and deflation techniques in a scenario where orthogonality does not hold and find that they still lead to significantly improved performance. 
\end{abstract}

\section{Introduction}
Principal Components Analysis (PCA, \citep{Hotelling:1933}) is a widely-used approach to finding low-dimensional patterns in complex data, enabling visualization, dimension reduction (compression), and predictive modeling. While PCA performs well in a wide range of low-dimensional settings, its performance degrades rapidly in high-dimensions, necessitating the use of regularized variants. Recently, \citet{Allen:2019} proposed Sparse and Functional PCA (SFPCA), a unified regularization scheme that allows for simultaneous smooth (functional) and sparse estimation of both row and column principal components (PCs). The rank-one SFPCA estimator is given by
\begin{equation}
\argmax_{\bu \in \overline{\B}^n_{\bSu}, \bv \in \overline{\B}^p_{\bSv}} \bu^T\bX\bv - \lambda_{\bu} P_{\bu}(\bu) - \lambda_{\bv} P_{\bv}(\bv) \label{eqn:sfpca}
\end{equation}
where $P_{\bu}(\cdot)$ is a regularizer inducing sparsity in the row PCs, with strength controlled by $\lambda_{\bu}$; $\bOmega_{\bu}$ is a positive semi-definite penalty matrix, typically a second- or fourth-order difference matrix; $\bSu = \bI + \alpha_{\bu} \bOmega_{\bu}$ is a smoothing matrix for the row PCs, with strength controlled by $\alpha_{\bu}$; and  $\overline{\B}^n_{\bSu}$ is the unit ellipse of the $\bSu$-norm, \emph{i.e.}, $\overline{\B}^n_{\bSu} = \{\bu \in \R^n: \bu^T\bSu\bu \leq 1\}$. (Respectively, $P_{\bv}(\cdot)$, $\lambda_{\bv}$, $\bOmega_{\bv}$, $\alpha_{\bv}$, $\bSv$, and $\overline{\B}^p_{\bSv}$ for the column PCs.)

\citet{Allen:2019} show that SFPCA unifies much of the existing regularized PCA literature \citep{Silverman:1996,Shen:2008,Huang:2008,Huang:2009,Witten:2009,Allen:2014} into a single framework, avoiding many pathologies of other approaches. Finally, they propose an efficient alternating maximization scheme with guaranteed global convergence to solve the bi-concave SFPCA problem \eqref{eqn:sfpca}. The SFPCA estimator only allows for a single pair of PCs to be estimated for a given data matrix $\bX$. \citeauthor{Allen:2019} suggest applying SFPCA repeatedly to the deflated data matrix if multiple PCs are desired. While this approach performs acceptably in practice, it loses the interpretable orthogonality properties of standard PCA. In particular, the estimated PCs are no longer guaranteed to be orthogonal to each other, hindering the common interpretation of PCs as statistically independent sources of variance, or to the deflated data matrix, suggesting that additional signal remains uncaptured.

We extend the work of \citet{Allen:2019} to address these shortcomings: first, in Section \ref{sec:man_opt}, we modify the SFPCA estimator to simultaneously estimate several sparse PCs subject to orthogonality constraints. The resulting estimator is constrained to a product of generalized Stiefel mainfolds and we propose three efficient algorithms to solve the resulting manifold optimization problem. Next, in Section \ref{sec:deflation}, we propose improved deflation schemes which provably remove all of the signal from the data matrix, allowing for more accurate iterative estimation of sparse PCs. Finally, we demonstrate the improved performance of our manifold estimators and deflation schemes in Section \ref{sec:sims}. Supplemental materials for this paper, including proofs, counter-examples, and additional algorithmic details, are available online at \url{https://arxiv.org/abs/1907.12012}.

\begin{table*}[ht]
\vspace{-0.13in}
\centering
\begin{tabular}{l|cccc}
\toprule
\multirow{2}{*}{\textbf{Method}} & Two-Way Orthogonality & One-Way Orthogonality & Subsequent Orthogonality ($\forall s \geq 0$) & Robust to\\
& $\bu_t^T\bX_t\bv_t = 0$ & $\bu_t^T\bX_t, \bX_t\bv_t = \bzero$ & $\bu_t^T\bX_{t + s}, \bX_{t + s}\bv_t = \bzero$ & Scale of $\bu_t, \bv_t$\\
\midrule
Hotelling's Deflation \eqref{eqn:hotelling_d} & \cmark & \xmark & \xmark & \xmark\\
Projection Deflation \eqref{eqn:projection_d}  & \cmark & \cmark & \xmark & \xmark \\
Schur Deflation \eqref{eqn:schur_d}  & \cmark & \cmark & \cmark & \cmark\\
\bottomrule
\end{tabular}
\caption{Properties of Hotelling's Deflation (HD), Projection Deflation (PD), and Schur Complement Deflation (SD). Only SD captures all of the individual signal of each principal component without re-introducing signal at later iterations. Additionally, only SD allows for the non-unit-norm PCs estimated by SFPCA to be used without rescaling.}
\label{tab:deflation_props}
\end{table*}

\section{Manifold Optimization for SFPCA} \label{sec:man_opt}

One of the most attractive properties of PCA is the factors it extracts are orthogonal ($\bu_t^T\bu_s = \bv_t^T\bv_s =1$ if $t = s$ and $0$ otherwise). Because of this, PCs can be interpreted as separate sources of variance and, under an additional Gaussianity assumption, statistically independent. While this follows directly from the properties of eigendecompositions for standard PCA, it is much more difficult to obtain similar results for sparse PCA. Some authors have suggested that there exists a fundamental tension between orthogonality and sparsity, with \citet{Journee:2010} calling the goal of sparse and orthogonal estimation ``questionable.'' Indeed \emph{ex post} orthogonalization of sparse PCs, \emph{e.g.}, using a Gram-Schmidt step, destroys any sparsity in the estimated PCs. To avoid this, it is necessary to impose orthogonality directly in the estimation step, rather than trying to impose it afterwards. 

We modify the SFPCA estimator to simultaneously estimate multiple PCs subject to an orthognality constraint:
\begin{equation}
\argmax_{\bU \in \V_{n\times k}^{\bSu}, \bV \in \V_{p \times k}^{\bSv}} \Tr(\bU^T\bX\bV) - \lambda_{\bU} P_{\bU}(\bU) - \lambda_{\bV} P_{\bV}(\bV) \label{eqn:mansfpca} 
\end{equation}
where $\V^{\bSu}_{n \times k}$ is the generalized Stiefel manifold of order $k$ over $\mathbb{R}^n$, \emph{i.e.},
\[\bU \in \V_{n \times k}^{\bSu} \Longleftrightarrow \bU \in \R^{n \times k} \text{ and } \bU^T\bSu\bU = \bI_k.\]
The generalized Stiefel manifold constraint ensures orthogonality of the estimated PCs, while still allowing us to capture most of the variability in the data.\footnote{When estimating orthogonal factors, it is common to re-express the problem using a (generalized) Grassmanian manifold constraint to avoid identifiability issues. We cannot use the Grassmanian approach here as sparse estimation (implicitly) fixes a single coordinate system.} We note that, because we use a generalized Stiefel constraint, the estimated PCs will be \emph{orthogonal with respect to $\bS_{\bu}$}, \emph{i.e.}, $\bu_t^T\bS_{\bu}\bu_s = 0$ for $t \neq s$, rather than orthogonal in the standard sense. This is commonly observed for functional PCA variants \citep{Silverman:1996,Huang:2008,Huang:2009} and can be interpreted as orthogonality under the inner product generating the $\bSu$-norm. If no roughness penalty is imposed ($\alpha_{\bu} = 0$ or $\alpha_{\bv} = 0$), then our method gives orthogonality in the standard sense.

To solve the Manifold SFPCA problem \eqref{eqn:mansfpca}, we employ an alternating maximization scheme, first holding $\bV$ fixed while we update $\bU$ and \emph{vice versa}, as described in Algorithm \ref{alg:man_sfpca_general}. Even with one parameter held fixed, the resulting sub-problems are still difficult manifold optimization problems, which require iterative approaches to obtain a solution \citep{Absil:2007,Wen:2013,Lai:2014,Kovantsky:2016,Chen:2018,Chen:2019,Benidis:2016,Sato:2019}.

\begin{algorithm}[H]
\caption{Manifold SFPCA Algorithm} \label{alg:man_sfpca_general}
\begin{enumerate}
  \item Initialize $\hat{\bU}, \hat{\bV}$ to the leading $k$ singular vectors of $\bX$
  \item Repeat until convergence:
  \begin{enumerate}[(a)]
    \item $\bU$-subproblem. Solve using Algorithm \ref{alg:man_sfpca_manpg} or \ref{alg:man_sfpca_madmm}:
    \[\hat{\bU} = \argmin_{\bU \in \V^{\bSu}_{n \times k}} -\Tr(\bU^T\bX\hat{\bV}) + \lambda_{\bU}P_{\bU}(\bU)\]
    \item $\bV$-subproblem: Solve using Algorithm \ref{alg:man_sfpca_manpg} or \ref{alg:man_sfpca_madmm}, with $\bU$ and $\bV$ reversed:
    \[\hat{\bV} = \argmin_{\bV \in \V^{\bSv}_{n \times p}} -\Tr(\hat{\bU}^T\bX\bV) + \lambda_{\bV}P_{\bV}(\bV)\]
  \end{enumerate}
  \item Return $\hat{\bU}$ and $\hat{\bV}$
\end{enumerate}
\end{algorithm}
\vspace{-0.1in}\noindent \citet{Allen:2019} developed a custom projected + proximal gradient algorithm to solve the $\bu$- and $\bv$ subproblems of the rank-one SFPCA estimator. Assuming $P_{\bU}$ and $P_{\bV}$ are positive homogeneous, (\emph{e.g.} $P(\cdot) = \|\bA\cdot\|_p$ for arbitrary $p \geq 1$ and $\bA$), they establish convergence to a stationary point. In order to extend this idea to the multi-rank (manifold) case, we use the recently-proposed Manifold Proximal Gradient (ManPG) scheme of \citet{Chen:2018}, detailed in Algorithm \ref{alg:man_sfpca_manpg}. ManPG proceeds in two-steps: first, we solve for a descent direction $\bD$ of the objective along the tangent space of the generalized Stiefel manifold, subject to the tangency constraint of $\bD^T\bSu\bU^{(k)}$ being skew-symmetric; secondly, back-tracking line search is used to determine a step-size $\alpha$, after which the estimate is projected back onto the generalized Steifel manifold using a retraction anchored at the previous $\bU^{(k)}$. The retraction, which plays the same role as the projection step in the original SFPCA algorithm \citep[Algorithm 1]{Allen:2019}, can be computed using a Cholesky factorization \citep[Algorithm 3.1]{Sato:2019}. \citet{Chen:2019} showed that a single step of ManPG is sufficient to ensure convergence despite the bi-concave objective: we refer to their approach as Alternating ManPG (A-ManPG).

\begin{algorithm}[H]
\caption{Manifold Prox.~Gradient (ManPG) for $\hat{\bU}$-Subproblem} \label{alg:man_sfpca_manpg}
\begin{enumerate}
  \item Initialize $\bU^{(k)} = \hat{\bU}$
  \item Repeat until convergence:
  \begin{itemize}
    \item Solve, subject to $\bD^T\bSu\bU^{(k)} + (\bU^{(k)})^T\bSu\bD = \bzero$:
    \[\hat{\bD} = \argmin_{\substack{\bD \in \R^{n \times k}}} -\langle \bX\hat{\bV}, \bD \rangle_F + \lambda_{\bU}P_{\bU}(\bU^{(k)} + \bD)\]
    \item Select $\alpha$ by Armijo-back-tracking
    \item $\bU^{(k+1)} = \textsf{Retr}_{\bU^{(k)}}(\alpha \hat{\bD})$
  \end{itemize}
  \item Return $\hat{\bU}$
\end{enumerate}
\end{algorithm}

\noindent Note that in general manifold proximal gradient schemes \citep{Chen:2018,Chen:2019} impose a maximum step-size to ensure that linearization of the smooth portion of the objective actually leads to descent: because the smooth portion of our objective function in linear in $\bU$ and $\bV$, we can omit this term from Algorithm \ref{alg:man_sfpca_manpg}.

While efficient when tuned properly, we have found the performance of ManPG on the $\bU$- and $\bV$-subproblems quite sensitive to infeasibility in the descent direction. A more robust scheme can be derived by using the Manifold ADMM (MADMM) scheme of \citet{Kovantsky:2016} to solve the subproblems. Like standard ADMM schemes, MADMM allows us to split a problem into two parts, each of which can be solved more easily than the global problem. When applied to the $\bU$- and $\bV$-subproblems, the MADMM allows us to separate the manifold constraint from the sparsity inducing regularizer, thereby side-stepping the orthogonality / sparsity tension at the heart of this paper. After this splitting, the smooth update can be shown to equivalent to the unbalanced Procrustes problem \citep{Schonemann:1966,Elden:1999} with a closed-form update: $\bU^{(k+1)} = \bSu^{-1/2}\bA\bB^T$ where $\bA\bDelta\bB^T$ is the SVD of $\bSu^{-1/2}\bX\hat{\bV} + \rho\bSu^{1/2}(\bW^{(k)} - \bZ^{(k)})$. The sparse update is simply the proximal operator of $P_{\bU}(\cdot)$, typically a threhsolding step \citep[Chapter 6]{Beck:2017}. To the best of our knowledge, the convergence of MADMM has not yet been established, but we have not observed significant non-convergence problems in our experiments.

\begin{algorithm}[H]
\caption{Manifold ADMM (MADMM) for $\hat{\bU}$-Subproblem} \label{alg:man_sfpca_madmm}
\begin{enumerate}
  \item Initialize $\bU^{(k)} = \bW^{(k)} = \hat{\bU}$, $\bZ^{(k)} = \bzero$ and $k = 1$
  \item Repeat until convergence:
  {\footnotesize
  \begin{align*}
  \bU^{(k+1)} &= \argmin_{\bU \in \V^{\bSu}_{n \times k}} - \Tr(\bU^T\bX\bV) + \frac{\rho}{2}\|\bU - \bW^{(k)} + \bZ^{(k)}\|_F^2 \\
  \bW^{(k+1)} &= \argmin_{\bW \in \R^{n \times k}} \lambda_{\bU}P_{\bU}(\bW) + \frac{\rho}{2}\|\bU^{(k+1)} - \bW + \bZ^{(k)}\|_F^2 \\
              &= \prox_{\lambda_{\bU}/\rho P_{\bU}(\cdot)}\left(\bU^{(k+1)} + \bZ^{(k)}\right) \\
  \bZ^{(k+1)} &= \bZ^{(k)} + \bU^{(k+1)} - \bW^{(k+1)}
\end{align*}}
  \item Return $\hat{\bU}$ and $\hat{\bV}$
\end{enumerate}
\end{algorithm}

\section{Iterative Deflation for SFPCA} \label{sec:deflation}
We next consider the use of iterative deflation schemes for multi-rank SFPCA. As discussed by \citet{Mackey:2008}, the attractive orthogonality properties of standard (Hotelling's) deflation depend critically on the estimated PCs being exact eigenvectors of the covariance matrix. Because the PCs estimated by sparse PCA schemes are almost surely not eigenvectors, \citet{Mackey:2008} proposes several alternate deflation schemes which retain some of the attractive properties of Hotelling's deflation even when non-eigenvectors are used. We extend these to the low-rank model and allow for deflation by several PCs, possibly with non-unit norm, simultaneously, \emph{e.g.}, as produced by ManSFPCA. To ease exposition, we first work in the vector setting and consider the general case at the end of this section. The properties of our proposed deflation schemes are summarized in Table \ref{tab:deflation_props} above.

The simplest deflation scheme is essentially that used by \citet{Hotelling:1933}, extended to the low-rank model:
\begin{equation}
  \bX_t \coloneqq \bX_{t-1} - d_t\bu_t\bv_t^T \text{ where } d_t = \bu_t^T\bX_{t-1}\bv_t. \tag{HD} \label{eqn:hotelling_d}
\end{equation}
For two-way sparse PCA variants \citep{Witten:2009,Allen:2014}, this deflation gives a deflated matrix which is ``two-way'' orthogonal to the estimated PCs, \emph{i.e.}, $\bu_t^T\bX_t\bv_t = 0$. We may interpret this as Hotelling's deflation (HD) capturing all the signal \emph{jointly} associated with the pair $(\bu_t, \bv_t)$.

We may also ask if HD captures \emph{all} of the signal associated with $\bu_t$ or only the signal which is also associated with $\bv_t$. If HD captures all of the signal associated with $\bu_t$, then we would expect $\bu_t^T\bX_t \tilde{\bv} = 0$ for all $\tilde{\bv} \in \R^p$, or equivalently, $\bu_t^T\bX_t = \bzero_p$. Interestingly, HD does not have this left-orthogonality property, suggesting that it leaves additional $\bu_t$-signal in the deflated matrix $\bX_t$. 

HD fails to yield left- and right-orthogonality because it is not based on a projection operator. To address this in the covariance model, \citet{Mackey:2008} proposed a deflation scheme which projects the covariance matrix onto the orthogonal complement of the estimated principal component. We extend this idea to the low-rank model by projecting the column- and row-space of the data matrix into the orthogonal complement of the left- and right-PCs respectively, giving two-way projection deflation (PD):
\begin{equation}
  \bX_t \coloneqq (\bI_n - \bu_t\bu_t^T)\bX_{t-1}(\bI_p - \bv_t\bv_t^T). \tag{PD} \label{eqn:projection_d}
\end{equation}
Unlike HD, PD captures all of the linear signal associated with $\bu_t$ and $\bv_t$ individually ($\bu_t\bX_t\tilde{\bv} = \tilde{\bu}\bX_t\bv_t = 0, \forall \tilde{\bu} \in \R^n, \tilde{\bv} \in \R^p$).

If we use PD repeatedly, however, the multiply-deflated matrix will not continue to be orthogonal to the PCs: that is, $\bu_t^T\bX_{t+s} \neq \bzero$ for $s \geq 1$. This suggests that repeated application of PD can \emph{reintroduce} signal in the direction of the PCs that we previously removed. This occurs because PD works by sequentially projecting the data matrix, but in general the compositition of two orthogonal projections is not another orthogonal projection without additional assumptions. To address this, \citet{Mackey:2008} proposed a Schur complement deflation (SD) technique, which we now extend to the low-rank (two-way) model:
\begin{equation}
  \bX_t \coloneqq \bX_{t-1} - \frac{\bX_{t-1}\bv_t\bu_t^T\bX_{t-1}}{\bu_t^T\bX_{t-1}\bv_t}. \tag{SD} \label{eqn:schur_d}
\end{equation}
While \citeauthor{Mackey:2008} motivates this approach using conditional distributions and a Gaussianity assumption on $\bX$, it can also be understood as an alternate projection construct which is more robust to scaling and non-orthogonality. 

So far, we have only considered the behavior of the proposed deflation schemes for two-way sparse PCA. If we consider SFPCA in generality, however, $\bu_t$ and $\bv_t$ are unit vectors under the $\bSu$- and $\bSv$-norms, not under the Euclidean norm. Consequently, the projections used by PD may not be actual projections and a PD deflated matrix may fail to be two- or one-way orthogonal. Normalizing the estimated PCs before deflation addresses this problem and is recommended in practice: conversely, because its deflation term is invariant under rescalings of $\bu_t$ and $\bv_t$, SD works without renormalization.

\begin{figure*}
\vspace{-0.13in}
\centering
\begin{tabular}{ccc|ccc}
\multicolumn{3}{c}{\it Scenario 1: $\bU^*$ and $\bV^*$ Orthogonal - SNR $\approx$ 1.2} & \multicolumn{3}{c}{\it Scenario 2: $\bU^*$ and $\bV^*$ \textbf{Not} Orthogonal - SNR $\approx$ 1.7} \\
Signal & SVD & ManSFPCA & Signal & SVD & ManSFPCA \\
\includegraphics[width=1in,height=0.55in]{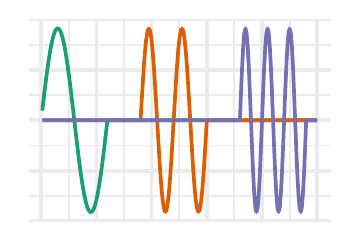} & \includegraphics[width=1in,height=0.55in]{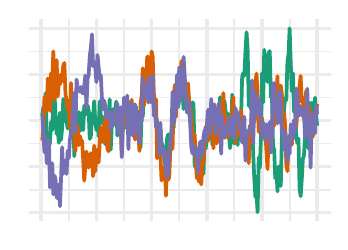} & \includegraphics[width=1in,height=0.55in]{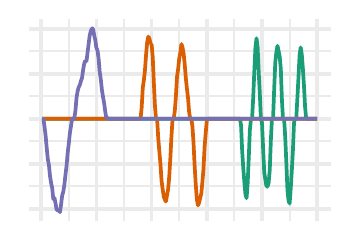} & \includegraphics[width=1in,height=0.55in]{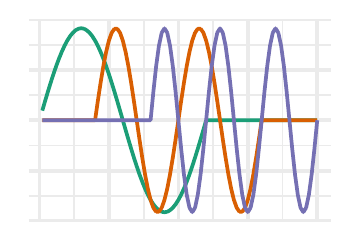} & \includegraphics[width=1in,height=0.55in]{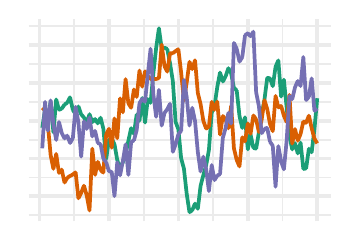} & \includegraphics[width=1in,height=0.55in]{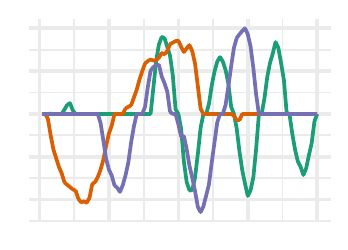} \\
\includegraphics[width=1in,height=0.55in]{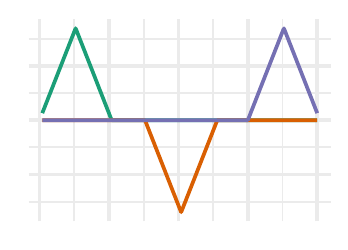} & \includegraphics[width=1in,height=0.55in]{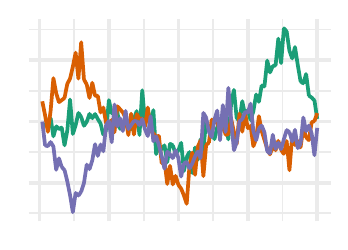} & \includegraphics[width=1in,height=0.55in]{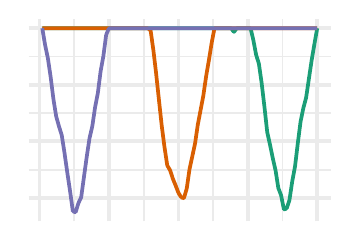} & \includegraphics[width=1in,height=0.55in]{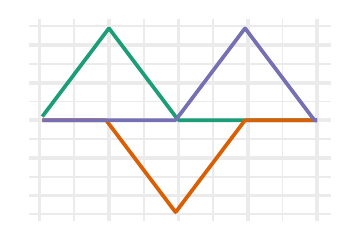} & \includegraphics[width=1in,height=0.55in]{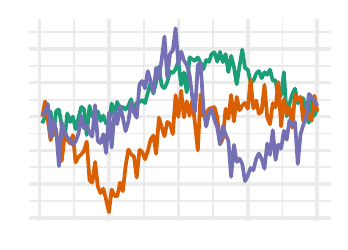} & \includegraphics[width=1in,height=0.55in]{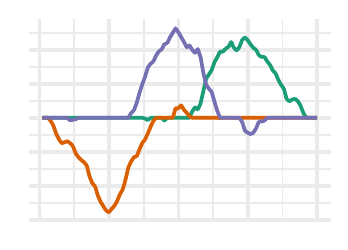} \\
\end{tabular}
\caption{Simulation Scenarios Used in Section \ref{sec:sims}. For both the left singular vectors (top row) and the right singular vectors (bottom), ManSFPCA is able to recover the signal far more accurately than unregularized PCA (SVD). ManSFPCA is not identifiable up to change of order or sign.}
\label{fig:sim_design}
\end{figure*}

The extension of these techniques to the multi-rank case is straightforward. We give the normalized variants here:
\begin{align*}
  \bX_t^{\text{HD}} &\coloneqq \bX_{t-1} - \bU_t(\bU_t^T\bU_t)^{-1}\bU_t^T\bX_{t-1}\bV_t(\bV_t^T\bV_t)^{-1}\bV_t^T\\
  \bX_t^{\text{PD}} &\coloneqq (\bI_n - \bU_t(\bU_t^T\bU_t)^{-1}\bU_t^T)\bX_{t-1}(\bI_p - \bV_t(\bV_t^T\bV_t)^{-1}\bV_t^T)\\
  \bX_t^{\text{SD}} &\coloneqq \bX_{t-1} - \bX_{t-1}\bV_t(\bU_t^T\bX_{t-1}\bV_t)^{-1}\bU_t^T\bX_{t-1}.
\end{align*}
As in the covariance model, if $\bu_t$ and $\bv_t$ are true singular vectors, all three deflation schemes are equivalent.

\section{Simulation Studies} \label{sec:sims}
In this section, we compare the performance of Manifold SFPCA and the iterative rank-one deflation schemes proposed above in illustrative simulation studies. Manifold SFPCA using Manifold ADMM (Algorithm \ref{alg:man_sfpca_madmm}) to solve the subproblems achieves better solutions than Manifold Proximal Gradient or A-ManPG (Algorithm \ref{alg:man_sfpca_manpg}) in less time. Furthermore, despite the additional flexibility of the iterative rank-one variants, Manifold SFPCA achieves both better signal recovery and a higher proportion of variance explained, even when the orthogonality assumptions are violated.

We first consider the relative performance of the three algorithms proposed for solving the Manifold SFPCA problem \eqref{eqn:mansfpca}. We generate data in $\bX^* = \bU^*\bD^*\bV^* \in \R^{250 \times 100}$ with three distinct PCs: the left PCs ($\bU^*$) are localized sinusoids of varying frequency; the right PCs ($\bV^*$) are non-overlapping sawtooth waves. (See Figure \ref{fig:sim_design}.) We add independent standard Gaussian noise ($\bE$) to give a signal-to-noise ratio (SNR) of $\|\bX^*\| / \|\bE\| \approx 1.2$.  We fix $\lambda_{\bu} = \lambda_{\bv} = 1$ and $\alpha_{\bu} = \alpha_{\bv} = 3$ which is near optimal for all three schemes. This is a favorable setting for Manifold SFPCA as the underlying signals are orthogonal, sparse, smooth, and of comparable magnitude.

Table \ref{tab:sim1} shows the performance of our three Manifold SFPCA algorithms on several metrics, averaged over 100 replicates. Overall, the Manifold ADMM (MADMM \citep{Kovantsky:2016}) and Alternating Manifold Proximal Gradient (A-ManPG \citep{Chen:2019}) variants perform best, handily beating the Manifold Proximal Gradient scheme \citep{Chen:2018} on all measures. MADMM achieved the best objective value on every replicate.

In terms of signal recovery, MADMM achieves slightly better performance than A-ManPG on the right singular vectors, while A-ManPG is slightly better on the left singular vectors. Computationally, MADMM dominates both proximal gradient variants even though it requires many more matrix decompositions.
The descent direction subproblems of ManPG and A-ManPG are rather expensive to solve repeatedly and their performance is very sensitive to the solver used. Overall, the MADMM variant of Manifold SFPCA achieves the best optimization and statistical performance in far less time than the proximal gradient-based variants.

Next, we compare Manifold SFPCA with the iterative deflation schemes proposed in Section \ref{sec:deflation} in two different scenarios: the favorable scenario used above and a less-favorable scenario where the true PCs are shifted and no longer orthogonal. ($n = p = 100$ and $\|(\bU^*)^T(\bU^*) - \bI_3\|, \|(\bV^*)^T\bV^* - \bI_3\| \approx 0.37$.) We compare the proportion of variance explained using Manifold SFPCA with iterative rank-one SFPCA using the normalized Hotelling, Projection, and Schur Complement deflation strategies. As can be seen in Table \ref{tab:sim2}, PD and SD consistently dominate HD. Because PD and SD fully remove the signal associated with estimated PCs, the subsequent PCs are able to capture different signals and explain a larger fraction of variance. By ensuring that the signal is never re-introduced, SD does even better than PD as we consider higher ranks. Interestingly, while PD and SD perform about as well when the underlying signals are orthogonal, SD performs much better in the non-orthogonal scenario. By estimating all three PCs simultaneously, Manifold SFPCA is able to find a better set of PCs than any of the greedy deflation methods.

\begin{table}[htb]
\centering
\begin{tabular}{ll|ccc}
\toprule
&& MADMM & ManPG & A-ManPG\\
\midrule
\multicolumn{2}{c|}{Time (s)} & \bf{19.46} & 217.40 & 175.99\\
\multicolumn{2}{c|}{Suboptimality} & \bf{0} & 4.58 & 12.23 \\
\midrule
\multirow{2}{*}{rSS-Error} & $\bU$ & 68.66\% & 74.03\% & \textbf{64.54\%} \\
& $\bV$ & \textbf{36.85\%} & 50.69\% & 43.17\%\\
\midrule
\multirow{2}{*}{TPR} & $\bU$ & 87.75\% & 69.72\% & \textbf{89.99\%} \\
& $\bV$ & \textbf{94.75\%} & 70.30\% & 89.87\%\\
\midrule
\multirow{2}{*}{FPR} & $\bU$ & 12.25\% & 30.28\% & \textbf{10.01\%} \\
& $\bV$ & \textbf{5.25\%} & 29.70\% & 10.13\% \\
\bottomrule
\end{tabular}
\caption{Comparison of Manifold-ADMM, Manifold Proximal Gradient, and Alternating Manifold Proximal Gradient approaches for Manifold SFPCA \eqref{eqn:mansfpca}. MADMM is consistently more efficient and obtains better solutions, but both MADMM and A-ManPG perform well in terms of signal recovery, as measured by relative subspace recovery error ($\text{rSS-Error} = \|\hat{\bU}\hat{\bU}^T - \bU^*(\bU^*)^T\| / \|\hat{\bU}_{\text{SVD}}\hat{\bU}_{\text{SVD}}^T - \bU^*(\bU^*)^T\|$), true positive rate (TPR) and false positive rate (FPR).}
\label{tab:sim1}
\end{table}

\begin{table}[htb]
\centering
\begin{tabular}{ll|cccc}
\toprule
\multicolumn{6}{c}{\it Scenario 1: $\bU^*$ and $\bV^*$ Orthogonal} \\
&& HD & PD & SD & ManSFPCA \\
\midrule
\multirow{3}{*}{CPVE} & PC1 & 15.92\% & 21.05\% & \textbf{21.87\%} & \multirow{3}{*}{\bf 37.12\%}\\
& PC2 & 22.21\% & 29.42\% & \textbf{30.59\%} \\
& PC3 & 26.80\% & 35.57\% & \textbf{37.09\%} \\
\midrule
\multicolumn{6}{c}{\it Scenario 2: $\bU^*$ and $\bV^*$ \textbf{Not}  Orthogonal} \\
\midrule
\multirow{3}{*}{CPVE} & PC1 & 8.85\% & 19.74\% & \textbf{29.80\%} & \multirow{3}{*}{\bf 50.85\%}\\
& PC2 & 13.03\% & 28.30\% & {\bf 39.87\%}\\
& PC3 & 16.16\% & 34.22\% & {\bf 46.48\%} \\
\bottomrule
\end{tabular}
\caption{Cumulative Proportion of Variance Explained (CPVE) of Rank-One SFPCA with (normalized) Hotelling, Projection, and Schur Complement Deflation and of (order 3) Manifold SFPCA. SD gives the best CVPE of the iterative approaches and appears to be more robust to violations of orthogonality (Scenario 2). Manifold SFPCA outperforms the iterative methods in both scenarios.}
\label{tab:sim2}
\end{table}

\section{Discussion}

We have introduced two practical extensions to Sparse and Functional PCA: first, we presented a multi-rank scheme which estimates multiple PCs simultaneously and proposed algorithms to solve the resulting manifold optimization problems. The resulting estimator inherits many of the attractive properties of rank-one SFPCA and is, to the best of our knowledge, the first multi-rank PCA scheme for the low-rank model. ManSFPCA combines both the flexiblity and superior statistical performance of rank-one SFPCA with the superior interpretability of orthogonal (non-regularized) PCA.

Secondly, we re-considered the use of Hotelling's deflation, and developed two additional deflation schemes which have attractive theoretical properties and emprical performance. Our schemes extend the results of \citet{Mackey:2008} in several ways: they allow for deflation by multiple PCs in a single step, they are applicable to the low-rank model and the covariance model, and they are robust to non-orthogonality and non-unit-scaling common to Functional PCA variants. While developed for SFPCA, these deflation schemes are useful for any regularized PCA model.

We note here that our results can be extended naturally to other multivariate analysis techniques which can be expressed in a regularized SVD framework (\emph{e.g.}, PLS, CCA, \emph{etc.}). Our deflation approaches can be also extended to the higher-order / multi-way context and may be particularly useful in the context of regularized CP decompositions \citep{Allen:2012,Allen:2013}. In the tensor setting, ManSFPCA is a sparse and smooth version of a Tucker decomposition, which suggests several interesting extensions we leave for future work \citep{Tucker:1966,Kolda:2009}.

\clearpage
\section{References}
\printbibliography[heading=none]
\end{refsection}

\begin{refsection}
\onecolumn
\appendix
\setcounter{algorithm}{0}
\renewcommand{\thealgorithm}{A\arabic{algorithm}}

\section*{Supplementary Materials}
\section{Proofs} \label{app:proofs}
\subsection{Deflation Schemes}  \label{app:proofs_deflation}
In this section, we give proofs of the claimed properties for the deflation schemes discussed in Section \ref{sec:deflation} and summarized in Table \ref{tab:deflation_props}.

We first consider Hotelling's deflation scheme \eqref{eqn:hotelling_d}: given estimated left- and right-singular vectors $\bu_t, \bv_t$ of a $n$-by-$p$ matrix $\bX_{t-1}$, two-way orthogonality is immediate:
\begin{align*}
  \bu_t^T\bX_t\bv_t &= \bu_t^T(\bX_{t-1} - d_t\bu_t\bv_t^T)\bv_t \\
                    &= \bu_t^T\bX_{t-1}\bv_t - d_t\bu^T\bu_t\bv_t^T\bv_t \\
                    &= d_t - d_t \|\bu_t\|^2 \|\bv_t\|^2
\end{align*}
If $\|\bu_t\|^2 = \|\bv_t\|^2 = 1$, then this gives the desired two-way orthogonality. From here, we clearly see that Hotelling's deflation only gives two-way orthogonality if $\bu_t$ and $\bv_t$ are unit-scaled, showing that it is not robust to non-unit scaling. Turning our attention to one-way orthogonality, we give a counter-example using the data matrix \[\bX = \begin{pmatrix*} 2 & -4/3 \\ 2 & \pp2/3 \\ 1 & \pp4/3 \end{pmatrix*} = \underbrace{\begin{pmatrix} 2/3 & -2/3 & \pp1/3 \\ 2/3 & \pp1/3 & -2/3 \\ 1/3 & \pp2/3 & \pp2/3 \end{pmatrix}}_{=\bU}\underbrace{\begin{pmatrix} 3 & 0 & 0 \\ 0 & 2 & 0 \\ 0 & 0 & 1\end{pmatrix}}_{=\bD} \underbrace{\begin{pmatrix} 1 & 0 & 0 \\ 0 & 1 & 0 \end{pmatrix}^T}_{= \bV^T}.\]
If we take take $\bu_1 = \begin{pmatrix} 1/\sqrt{2} & 1/\sqrt{2} & 0 \end{pmatrix}^T$ as a sparse left singular vector and $\bv_1 = \begin{pmatrix} 1 & 0 \end{pmatrix}^T$ the true right singular vector, then Hotelling's deflation gives $d_1 = \sqrt{8}$ and \[\bX_1 = \begin{pmatrix} 0 & -4/3 \\ 0 & \pp2/3 \\ 1 & \pp4/3 \end{pmatrix}.\] The deflated $\bX_1$ is not left-orthogonal to $\bu_1$, however, as $\bu_1^T\bX_1 = \begin{pmatrix} 0 & -\sqrt{2} / 3 \end{pmatrix}$, showing that Hotelling's deflation is not one-way orthogonal.

Next we consider the projection deflation scheme \eqref{eqn:projection_d}. Left orthogonality can be shown explicitly:
\begin{align*}
  \bu_t^T\bX_t &= \bu^T_t(\bI_n - \bu_t\bu_t^T)\bX_{t-1}(\bI_p - \bv_t\bv_t^T) \\
  &= (\bu_t^T - \|\bu_t\|^2\bu_t^T)\bX_{t-1}(\bI_p - \bv_t\bv_t^T)
\end{align*}
which is clearly zero for arbitrary $\bv_t$ if and only if $\|\bu_t\| = 1$. Essentially the same argument shows that projection deflation is right orthogonal if and only if $\|\bv_t\| = 1$, and two-way orthogonality follows from either left or right orthogonality. Hence projection deflation is both one- and two-way orthogonal but it is also sensitive to the unit scaling of the left and right principal components. Turning to subsequent orthogonality, we now take 
\[\bX = \begin{pmatrix*}[r] -2 & -3/2 & 1 \\ 8/3 & 1/6 & 1/3 \\ 0 & 5/2 & 1 \\ 2/3 & 7/6 & 7/3 \end{pmatrix*} = \underbrace{\begin{pmatrix*}[r] -1/2 & 1/2 & 1/2 & 1/2 \\ 1/2 & -1/2 & 1/2 & 1/2 \\ 1/2 & 1/2 & -1/2 & 1/2 \\ 1/2 & 1/2 & 1/2 & -1/2\end{pmatrix*}}_{=\bU} \underbrace{\begin{pmatrix*}[r] 4 & 0 & 0 & 0 \\ 0 & 3 & 0 & 0 \\ 0 & 0 & 2 & 0 \\ 0 & 0 & 0 & 1 \end{pmatrix*}}_{=\bD} \underbrace{\begin{pmatrix*}[r] 2/3 & 2/3 & 1/3 \\ - 2/3 & 1/3 & 2/3 \\ 1/3 & -2/3 & 2/3 \\ 0 & 0 & 0 \end{pmatrix*}}_{=\bV^T}.\]
We take $\bu_1 = \begin{pmatrix} 1/2 & 1/2 & 1/2 & 1/2 \end{pmatrix}^T$ and $\bv_1 = \begin{pmatrix} 1/\sqrt{2} & 1/\sqrt{2} & 0 \end{pmatrix}^T$ as a pair of sparse PCs to yield the deflated matrix: 
\[\bX_1 = \begin{pmatrix*}[r] -1 /8 & 1/8 & -1/6 \\ 11/8 & -11/8 &-5/6 \\ -9/8 & 9/8 & -1/6\\ -1/8 & 1/8 & 7/6 \end{pmatrix*}\] The leading singular pair of $\bX_1$ are approximately $\begin{pmatrix} -0.0447 & -0.1673 & 0.8620 & 0.4765 \end{pmatrix}^T$ and $\begin{pmatrix} 0.6731 & 0.2168 & -0.7071 \end{pmatrix}^T$, which we sparsely approximate by $\bu_2 = \begin{pmatrix} 0 & 0 & 4/5 & 3/5 \end{pmatrix}^T$ and $\bv_2 = \begin{pmatrix} 1/\sqrt{2} & 0 & 1/\sqrt{2} \end{pmatrix}$ respectively. Another round of PD gives \[\bX_2 \approx \begin{pmatrix*}[r] 0.0417 & 0.2917 & 0 \\ 2.2083 & -0.5417 & 0 \\ 0.2750 & 0.9650 & 0 \\ -0.3667 & -1.2867 & 0 \end{pmatrix*}\] which has $\bu_1^T\bX_2 \approx \begin{pmatrix} 1.0792 & -0.2858 & 0 \end{pmatrix}$ which is clearly non-zero. 

Finally, we consider Schur complement deflation \eqref{eqn:schur_d}. As with projection deflation, left- and right-orthogonality (and hence two-way orthogonality), can be shown explicitly:
\begin{align*}
  \bu_t^T\bX_t &= \bu_t^T\left(\bX_{t-1} - \frac{\bX_{t-1}\bv_t\bu_t^T\bX_{t-1}}{\bu_t^T\bX_{t-1}\bv_t}\right) \\
  &=\bu_t^T\bX_{t-1} - \frac{(\bu_t^T\bX_{t-1}\bv_t)\bu_t^T\bX_{t-1}}{\bu_t^T\bX_{t-1}\bv_t} \\
  &= \bzero
\end{align*}
which holds without any further restrictions on $\bu_t$ or $\bv_t$, showing that Schur complement deflation is robust to unit scaling of estimated principal components. Subsequent left orthogonality can be shown by induction using the above result as the base case: for the inductive step, suppose $\bu_t^T\bX_{t + s} = \bzero$ for some $s$, then
\begin{align*}
  \bu_t^T\bX_{t + s + 1} &= \bu_t^T\left(\bX_{t+s} - \frac{\bX_{t+s}\bv_{t+s + 1}\bu_{t+s + 1}\bX_{t+s}}{\bu_{t+s+1}^T\bX_{t+s}\bv_{t+s+1}}\right) \\
  &= \bu_t^T\bX_{t+s} - \frac{(\bu_t^T\bX_{t+s})\bv_{t+s + 1}\bu_{t+s + 1}\bX_{t+s}}{\bu_{t+s+1}^T\bX_{t+s}\bv_{t+s+1}} \\
  &= \bzero - \frac{\bzero\bv_{t+s + 1}\bu_{t+s + 1}\bX_{t+s}}{\bu_{t+s+1}^T\bX_{t+s}\bv_{t+s+1}} \\
  &= \bzero
\end{align*}
as desired.

\subsection{Simultaneous Multi-Rank Deflation}
In this section, we extend the results of the previous section to allow for simultaneous deflation by several principal components at once. The properties of the multi-rank deflation schemes are essentially the same as those of their counterparts. The normalized decompositions are given by: 
\begin{align*}
  \bX_t^{\text{HD}} &\coloneqq \bX_{t-1} - \bU_t(\bU_t^T\bU_t)^{-1}\bU_t^T\bX_{t-1}\bV_t(\bV_t^T\bV_t)^{-1}\bV_t^T\\
  \bX_t^{\text{PD}} &\coloneqq (\bI_n - \bU_t(\bU_t^T\bU_t)^{-1}\bU_t^T)\bX_{t-1}(\bI_p - \bV_t(\bV_t^T\bV_t)^{-1}\bV_t^T)\\
  \bX_t^{\text{SD}} &\coloneqq \bX_{t-1} - \bX_{t-1}\bV_t(\bU_t^T\bX_{t-1}\bV_t)^{-1}\bU_t^T\bX_{t-1}.
\end{align*}
As before, we see that Hotelling's deflation gives two-way orthogonality: 
\begin{align*}
  \bU_t^T\bX_t^{\text{HD}}\bV_t &= \bU_t^T\left[\bX_{t-1} -  \bU_t(\bU_t^T\bU_t)^{-1}\bU_t^T\bX_{t-1}\bV_t(\bV_t^T\bV_t)^{-1}\bV_t^T\right]\bV_t \\
  &= \bU_t^T\bX_{t-1}\bV_t - \bU_t^T\bU_t(\bU_t^T\bU_t)^{-1}\bU_t\bX_{t-1}\bV_t(\bV_t^T\bV_t)^{-1}\bV_t^T\bV_t\\
  &= \bU_t^T\bX_{t-1}\bV_t - \bU_t^T\bX_{t-1}\bV_t \\
  &= \bzero.
\end{align*}
We do not get one-way orthogonality unless $\bV_t(\bV_t^T\bV_t)^{-1}\bV_t^T = \bI$:
\begin{align*}
  \bU_t^T\bX_t^{\text{HD}} &= \bU_t^T\left[\bX_{t-1} - \bU_t(\bU_t^T\bU_t)^{-1}\bU_t^T\bX_{t-1}\bV_t(\bV_t^T\bV_t)^{-1}\bV_t^T\right] \\
  &= \bU_t^T\bX_{t-1} - \bU_t^T\bU_t(\bU_t^T\bU_t)^{-1}\bU_t\bX_{t-1}\bV_t(\bV_t^T\bV_t)^{-1}\bV_t^T\\
  &= \bU_t^T\bX_{t-1} - \bU_t^T\bX_{t-1}\bV_t(\bV_t^T\bV_t)^{-1}\bV_t^T. 
\end{align*}
For projection deflation, we prove (left) one-way orthogonality as before: 
\begin{align*}
  \bU_t^T\bX_t^{\text{PD}} &= \bU_t^T\left[(\bI_n - \bU_t(\bU_t^T\bU_t)^{-1}\bU_t^T)\bX_{t-1}(\bI_p - \bV_t(\bV_t^T\bV_t)^{-1}\bV_t^T)\right] \\
  &=(\bU_t^T - \bU_t^T\bU_t(\bU_t^T\bU_t)^{-1}\bU_t^T)\bX_{t-1}(\bI_p - \bV_t(\bV_t^T\bV_t)^{-1}\bV_t^T) \\
  &= (\bU_t^T - \bU_t^T)\bX_{t-1}(\bI_p - \bV_t(\bV_t^T\bV_t)^{-1}\bV_t^T) \\
  &= \bzero.
\end{align*}
An essentially identical argument proves right one-way orthogonality. Either form of one-way orthogonality proves two-way orthogonality \emph{a fortiori}. Considering (left) subsequent orthogonality,
\begin{align*}
  \bU_t^T\bX_{t+1} &= \bU_t^T\left[(\bI - \bU_{t+1}(\bU_{t+1}^T\bU_{t+1})^{-1}\bU_{t+1}^T)\bX_t(\bI - \bV_{t+1}(\bV_{t+1}^T\bV_{t+1})^{-1}\bV_{t+1}^T)\right]
\end{align*}
we see that it cannot hold unless either $\bU_{t}$ lies in the (left) range-space of $\bU_{t+1}$ or commutes with $(\bI - \bU_{t+1}(\bU_{t+1}^T\bU_{t+1})^{-1}\bU_{t+1}^T)$, neither of which hold in general. Interestingly, the matrix form makes the necessity of normalization clear for HD and PD: the unnormalized form is essentially assuming $\bU_t^T\bU_t = \bI$, which will not hold unless we are in the non-smoothed case  ($\alpha_{\bu} = 0$). 

Finally, for Schur complement deflation, we get left-orthogonality via the same argument as the vector case: 
\begin{align*}
  \bU_t^T\bX_t^{\text{SD}} &= \bU_t^T\left[\bX_{t-1} - \bX_{t-1}\bV_t(\bU_t^T\bX_{t-1}\bV_t)^{-1}\bU_t^T\bX_{t-1}\right] \\
  &= \bU_t^T\bX_{t-1} - \bU_t^T\bX_{t-1}\bV_t(\bU_t^T\bX_{t-1}\bV_t)^{-1}\bU_t^T\bX_{t-1} \\
  &= \bU_t^T\bX_{t-1} - \bI \, \bU_t^T\bX_{t-1} \\
  &= \bzero.
\end{align*}
Taking this as the base case for subsequent orthogonality, we inductively assume $\bU_t^T\bX_{t+s} = \bzero$ and consider $\bU_t^T\bX_{t + s + 1}$:
\begin{align*}
  \bU_t^T\bX_{t+s+1} &= \bU_t^T\left[\bX_{t+s} - \bX_{t+s}\bV_{t+s+1}(\bU_{t+s+1}\bX_{t+s}\bV_{t+s+1})^{-1}\bU_{t+s+1}^T\bX_{t+s}\right] \\
  &= \bU_t^T\bX_{t+s} - \bU_t^T\bX_{t+s}\bV_{t+s+1}(\bU_{t+s+1}\bX_{t+s}\bV_{t+s+1})^{-1}\bU_{t+s+1}^T\bX_{t+s} \\
  &= \bzero - \bzero \, \bV_{t+s+1}(\bU_{t+s+1}\bX_{t+s}\bV_{t+s+1})^{-1}\bU_{t+s+1}^T\bX_{t+s} \\ 
  &= \bzero
\end{align*}
showing that SD indeed gives subsequent orthogonality. As with the vector case, SD is not sensitive to normalization, as indicated by the fact it has no $(\bU_t^T\bU_t)^{-1}$ or $(\bV_t^T\bV_t)^{-1}$ terms. 

\subsection{Solution of the Generalized Unbalanced Procrustes Problem}
\label{app:proof_gupp}
The $\bU$-update in the Manifold ADMM scheme for the $\hat{\bU}$-subproblem (Algorithm \ref{alg:man_sfpca_madmm}) requires us to solve the following problem:
  \[\hat{\bU} = \argmin_{\bU \in \V^{\bSu}_{n \times k}} -\Tr(\bU^T\bX\hat{\bV}) + \frac{\rho}{2} \|\bU - \bW^{(k)} + \bZ^{(k)}\|_{\bSu}^2\]
  which has the closed-form solution:
  \[\hat{\bU} = \bSu^{-1/2}\bA\bB^T \text{ where } \bA, \bDelta, \bB^T = \SVD(\bSu^{-1/2}\bX\hat{\bV} + \rho\bSu^{1/2}(\bW^{(k)} - \bZ^{(k)}))\]
  as shown by the following theorem.
\begin{theorem}
Suppose $\bS_{\bu}$ is a strictly positive definite $n\times n$ matrix and $\bA, \bB$ are full (column) rank matrices of size $n \times k$ for $k < n$. Then the solution to
  \[\hat{\bX} = \argmin_{\bX \in \V^{\bSu}_{n \times k}} -\Tr(\bX^T\bA) + \frac{\rho}{2} \|\bX - \bB\|_{\bSu}^2\]
 is given by
 \[\bX = \bSu^{-1/2}\bU\bV^T\]
 where $\bU\bD\bV^T$ is the (economical) SVD of $\bSu^{-1/2}\bA + \rho\bSu^{1/2}\bB$.
\end{theorem}

We note that this result is a (slight) generalization of the well-studied Procrustes problem first considered by \citet{Schonemann:1966} in the orthogonal case and extended to the unbalanced (Stiefel manifold) case by \citet{Elden:1999}. We modify their result to the generalized Stiefel manifold, though we do not consider the additional complexities associated with rank-deficient $\bA$, $\bB$, or $\bSu$ matrices as they do not apply to our problem. 

\begin{proof}
Let $\bY = \bSu^{1/2}\bX$ so that $\bX^T\bSu\bX \Leftrightarrow \bY^T\bY = \bI$.  Then we can rewrite the above problem as:
\begin{align*}
  \hat{\bX} &= \argmin_{\bX \in \V^{\bSu}_{n \times k}} -\Tr(\bX^T\bA) + \frac{\rho}{2} \|\bX - \bB\|_{\bSu}^2 \\
            &= \argmin_{\bX \in \V^{\bSu}_{n \times k}} -\Tr((\bSu^{1/2}\bX)^T(\bSu^{-1/2}\bA)) + \frac{\rho}{2} \|\bSu^{1/2}\bX - \bSu^{1/2}\bB\|_F^2 \\
  \hat{\bY} &= \argmin_{\bY \in \V_{n \times k}} -\Tr(\bY^T (\bSu^{-1/2}\bA)) + \frac{\rho}{2}\|\bY - \bSu^{1/2}\bB\|_F^2
\end{align*}
From here, we apply Lemma \ref{lem:procrustes_solution} to obtain:
\[\hat{\bY} = \bU\bV^T \text { where } \bU, \bD, \bV^T = \SVD(\bSu^{-1/2}\bA + \rho\bSu^{1/2}\bB)\]
and hence
\[\hat{\bX} = \bS_u^{-1/2}\bU\bV^T  \text { where } \bU, \bD, \bV^T = \SVD(\bSu^{-1/2}\bA + \rho\bSu^{1/2}\bB)\]
\end{proof}
\begin{lemma} \label{lem:procrustes_solution}
Suppose $\bA, \bB$ are full (column) rank matrices of size $n \times k$ for $k < n$. Then the solution to
  \[\hat{\bX} = \argmin_{\bX \in \V_{n \times k}} -\Tr(\bX^T\bA) + \frac{\rho}{2} \|\bX - \bB\|_{F}^2\]
 is given by
 \[\hat{\bX} = \bU\bV^T\]
 where $\bU\bD\bV^T$ is the (economical) SVD of $\bA + \rho \bB$.
\end{lemma}

\begin{proof}
  Letting $\langle A, B \rangle_F = \Tr(A^TB)$ be the standard (Frobenius) inner product, the above problem becomes:
  \begin{align*}
    \hat{\bX} &= \argmin_{\bX \in \V_{n \times k}} -\Tr(\bX^T\bA) + \frac{\rho}{2} \|\bX - \bB\|_{F}^2 \\
              &= \argmin_{\bX \in \V_{n \times k}} -\langle \bX, \bA \rangle_F + \frac{\rho}{2}\langle \bX - \bB, \bX - \bB \rangle_F \\
              &= \argmin_{\bX \in \V_{n \times k}} -\langle \bX, \bA \rangle_F + \frac{\rho}{2}\underbrace{\|\bX\|_F^2}_{=1}  -\rho \langle \bX,  \bB \rangle_F + \frac{\rho}{2}\underbrace{\|\bB\|_F^2}_{\text{constant}}\\
              &= \argmax_{\bX \in \V_{n \times k}} \langle \bX, \bA + \rho \bB \rangle_F
  \end{align*}
  Let $\bU\bD\bV^T$ be the SVD of $\bA + \rho \bB$. Then
  \[\langle \bX, \bA + \rho \bB \rangle_F = \langle \bX, \bU\bD\bV^T \rangle_F = \langle \bU^T\bX\bV, \bD \rangle_F\]
  Since $\bD$ is a diagonal matrix, this is maximized when the left term is an identity matrix, as can be obtained by taking $\bX = \bU\bV^T$.
\end{proof}

\section{Algorithmic Details}
In this section, we give additional details of the Algorithms used to solve the Manifold SFPCA \eqref{eqn:mansfpca} problem:
\begin{equation*}
\argmax_{\bU \in \V_{n\times k}^{\bSu}, \bV \in \V_{p \times k}^{\bSv}} \Tr(\bU^T\bX\bV) - \lambda_{\bU} P_{\bU}(\bU) - \lambda_{\bV} P_{\bV}(\bV) \end{equation*}

\subsection{Manifold Proximal Gradient}

Manifold proximal gradient \citep{Chen:2018,Chen:2019} proceeds in two steps: first, a descent direction within the tangent space is identified; secondly, a step along along the descent direction is taken, with the step size chosen by \citeauthor{Armijo:1966}-type \citep{Armijo:1966} back-tracking. (The line search method we use is essentially that of \citet{Beck:2010}; see also \citet[Section 4.2]{Parikh:2013}.) Because the step in the descent direction is taken in the ambient space rather than the along the manifold in question (\emph{e.g.}, an unconstrained step in $\R^{n\times k}$ rather than a geodesic move along $\V_{n \times k}$), a retraction step is used to project back onto the manifold and preserve feasibility.

Both steps require further discussion: we first consider identifying the descent direction, which requires solving the following problem
\[\hat{\bD}_{\bU} = \argmin_{\bD_{\bU} \in \R^{n \times k}} -\langle \bX\hat{\bV}, \bD_{\bU}\rangle_F + \lambda_{\bU}P_{\bU}(\bU^{(k)} + \bD_{\bU})\text{ subject to } \bD^T_{\bU}\bSu\bU^{(k)} + (\bU^{(k)})^T\bSu\bD_{\bU} = \bzero.\]
The constraint arises from the tangent space of the generalized Stiefel manifold \citep[Appendix A.1]{Chen:2019}, while the objective is essentially that of the overall problem. (Note that, because the smooth portion of the objective is already linear, we do not need to linearize it, unlike \citet{Chen:2018,Chen:2019}.) The tangency constraint makes this problem non-trivial to solve, but it can be reformulated as a linearly constrained quadratic program by splitting $\bD_{\bU}$ into positive and negative parts and solved using standard approaches. \citet{Chen:2018,Chen:2019} recommend the use of a semi-smooth Newton method to solve this problem \citep{Li:2016c,Ali:2017}, though we used the generic \textsf{SDPT3} solver of \citeauthor{Toh:1999} \citep{Toh:1999,Tutuncu:2003,Toh:2012} in our timing experiments.

Expanding Algorithm \ref{alg:man_sfpca_general} with Algorithm \ref{alg:man_sfpca_manpg} for the subproblems, we obtain Algorithm \ref{alg:man_sfpca_manpg_full} below. Because Algorithm \ref{alg:man_sfpca_manpg_full} is initialized at an infeasible pair $(\hat{\bU}, \hat{\bV})$ (unless $\alpha_{\bu} = \alpha_{\bv} = 0$) the first iteration almost always decreases the objective value: after that, however, each sub-problem typically increases the objective value. (If $\lambda_{\bu} = \lambda_{\bv} = 0$, this is guaranteed because the subproblems have a unique global optimum, but we cannot prove monotonicity in general.)

\citet{Chen:2019} suggested an interesting variant of this approach which requires only a single proximal gradient step at each iteration. Applying this approach to our problem yields Algorithm \ref{alg:man_sfpca_amanpg_full}. Somewhat surprisingly, they are able to give convergence guarantees for this approach, even though it has both non-convex constraints and a non-convex (bi-convex) objective. As far as we know, this is the only one of our algorithms to have provable convergence.

\begin{algorithm}[tb]
\caption{Alternating Maximization Approach for Manifold SFPCA \eqref{eqn:mansfpca} using Manifold Proximal Gradient for Subproblem Solutions} \label{alg:man_sfpca_manpg_full}
\begin{enumerate}
  \item Initialize $\hat{\bU}$ and $\hat{\bV}$ as the $k$ leading singular vectors of $\bX$
  \item Repeat Until Convergence:
  \begin{enumerate}
    \item Solve $\bU$-Subproblem using Manifold Proximal Gradient \citep{Chen:2018}:
    \[\hat{\bU} = \argmin_{\bU \in \V^{\bSu}_{n \times k}} -\Tr(\bU^T\bX\hat{\bV}) + \lambda_{\bU} P_{\bU}(\bU)\]
    \begin{enumerate}
      \item Initialize $\bU^{(k)} = \hat{\bU}$
      \item Repeat Until Convergence:
        \begin{enumerate}
          \item Determine Descent Direction:
            \[\hat{\bD}_{\bU} = \argmin_{\bD_{\bU} \in \R^{n \times k}} -\Tr(\bD_{\bU}^T\bX\hat{\bV}) + \lambda_{\bU}\|\bU^{(k)} + \bD_{\bU}\|_1 \text{ subject to } \bD_{\bU}^T\bSu\bU^{(k)} + (\bU^{(k)})^T\bSu\bD_{\bU} = \bzero\]
          \item Perform Backtracking to Determine Step Size:
          \begin{itemize}
            \item Set $\alpha = 1$
            \item While $-\Tr(\Retr_{\bU^{(k)}}(\alpha \bD_{\bU})^T\bX\hat{\bV}) + \lambda_{\bU}\|\Retr_{\bU^{(k)}}(\alpha \bD_{\bU})\|_1 > -\Tr((\bU^{(k)})^T\bX\hat{\bV}) + \lambda_{\bU}\|\bU^{(k)}\|_1$:
            \begin{itemize}
              \item Set $\alpha = 0.8 * \alpha$
            \end{itemize}
          \end{itemize}
          \item Set $\bU^{(k+1)} = \Retr_{\bU^{(k)}}(\alpha \bD_{\bU})$
      \end{enumerate}
      \item Set $\hat{\bU} = \bU^{(k)}$
    \end{enumerate}
    \item Solve $\bV$-Subproblem using Manifold Proximal Gradient \citep{Chen:2018}:
    \[\hat{\bV} = \argmin_{\bV \in \V^{\bSv}_{p \times k}} -\Tr(\hat{\bU}^T\bX\bV) + \lambda_{\bV} P_{\bV}(\bV)\]
    \begin{enumerate}
      \item Initialize $\bV^{(k)} = \hat{\bV}$
      \item Repeat Until Convergence:
        \begin{enumerate}
          \item Determine Descent Direction:
            \[\hat{\bD}_{\bV} = \argmin_{\bD_{\bV} \in \R^{p \times k}} -\Tr(\hat{\bU}^T\bX\bD_{\bV}) + \lambda_{\bV}\|\bV^{(k)} + \bD_{\bV}\|_1 \text{ subject to } \bD_{\bV}^T\bSv\bV^{(k)} + (\bV^{(k)})^T\bSv\bD_{\bV} = \bzero\]
          \item Perform Backtracking to Determine Step Size:
          \begin{itemize}
            \item Set $\alpha = 1$
            \item While $-\Tr(\hat{\bU}^T\bX\Retr_{\bV^{(k)}}(\alpha \bD_{\bU})) + \lambda_{\bV}\|\Retr_{\bV^{(k)}}(\alpha \bD_{\bV})\|_1 > -\Tr(\hat{\bU}\bX\bV^{(k)}) + \lambda_{\bV}\|\bV^{(k)}\|_1$:
            \begin{itemize}
              \item Set $\alpha = 0.8 * \alpha$
            \end{itemize}
          \end{itemize}
          \item Set $\bV^{(k+1)} = \Retr_{\bV^{(k)}}(\alpha \bD_{\bV})$
      \end{enumerate}
    \end{enumerate}
    \item Set $\hat{\bV} = \bV^{(k)}$
  \end{enumerate}
  \item Return $\hat{\bU}$ and $\hat{\bV}$
\end{enumerate}
\end{algorithm}

\begin{algorithm}[htb]
\caption{Alternating Manifold Proximal Gradient Approach \citep{Chen:2019} for Manifold SFPCA \eqref{eqn:mansfpca}}
\label{alg:man_sfpca_amanpg_full}
\begin{enumerate}
  \item Initialize $\hat{\bU}$ and $\hat{\bV}$ as the $k$ leading singular vectors of $\bX$
  \item Repeat Until Convergence:
  \begin{enumerate}
    \item $\bU$-Update: One Step of Manifold Proximal Gradient \citep{Chen:2018}:
    \[\hat{\bU} = \argmin_{\bU \in \V^{\bSu}_{n \times k}} -\Tr(\bU^T\bX\hat{\bV}) + \lambda_{\bU} P_{\bU}(\bU)\]
    \begin{enumerate}
      \item Determine Descent Direction:
        \[\hat{\bD}_{\bU} = \argmin_{\bD_{\bU} \in \R^{n \times k}} -\Tr(\bD_{\bU}^T\bX\hat{\bV}) + \lambda_{\bU}\|\hat{\bU} + \bD_{\bU}\|_1 \text{ subject to } \bD_{\bU}^T\bSu\hat{\bU} + \hat{\bU}^T\bSu\bD_{\bU} = \bzero\]
      \item Perform Backtracking to Determine Step Size:
        \begin{itemize}
          \item Set $\alpha = 1$
          \item While $-\Tr(\Retr_{\hat{\bU}}(\alpha \bD_{\bU})^T\bX\hat{\bV}) + \lambda_{\bU}\|\Retr_{\hat{\bU}}(\alpha \bD_{\bU})\|_1 > -\Tr(\hat{\bU}^T\bX\hat{\bV}) + \lambda_{\bU}\|\hat{\bU}\|_1$:
            \begin{itemize}
              \item Set $\alpha = 0.8 * \alpha$
            \end{itemize}
          \end{itemize}
          \item Set $\hat{\bU} = \Retr_{\hat{\bU}}(\alpha \bD_{\bU})$
    \end{enumerate}
    \item $\bV$-Update: One Step of Manifold Proximal Gradient \citep{Chen:2018}:
    \[\hat{\bV} = \argmin_{\bV \in \V^{\bSv}_{p \times k}} -\Tr(\hat{\bU}^T\bX\bV) + \lambda_{\bV} P_{\bV}(\bV)\]
    \begin{enumerate}
      \item Determine Descent Direction:
        \[\hat{\bD}_{\bV} = \argmin_{\bD_{\bV} \in \R^{p \times k}} -\Tr(\hat{\bU}^T\bX\bD_{\bV}) + \lambda_{\bV}\|\hat{\bV} + \bD_{\bV}\|_1 \text{ subject to } \bD_{\bV}^T\bSv\hat{\bV} + \hat{\bV}^T\bSv\bD_{\bV} = \bzero\]
      \item Perform Backtracking to Determine Step Size:
        \begin{itemize}
          \item Set $\alpha = 1$
          \item While $-\Tr(\hat{\bU}^T\bX\Retr_{\hat{\bV}}(\alpha \bD_{\bV})) + \lambda_{\bV}\|\Retr_{\hat{\bV}}(\alpha \bD_{\bV})\|_1 > -\Tr(\hat{\bU}^T\bX\hat{\bV}) + \lambda_{\bV}\|\hat{\bV}\|_1$:
            \begin{itemize}
              \item Set $\alpha = 0.8 * \alpha$
            \end{itemize}
          \end{itemize}
          \item Set $\hat{\bV} = \Retr_{\hat{\bV}}(\alpha \bD_{\bV})$
    \end{enumerate}
  \end{enumerate}
  \item Return $\hat{\bU}$ and $\hat{\bV}$
\end{enumerate}
\end{algorithm}

\subsection{Manifold ADMM}

If we expand Algorithm \ref{alg:man_sfpca_general} using Algorithm \ref{alg:man_sfpca_madmm} (Manifold ADMM \citep{Kovantsky:2016}) to solve the sub-problems, we obtain Algorithm \ref{alg:man_sfpca_madmm_full}.
Note that the result of Section \ref{app:proof_gupp} is used to solve the $\bU^{(k+1)}$ and $\bV^{(k+1)}$ updates appearing in Steps 2(a)(ii) and 2(b)(ii) respectively. Because Algorithm \ref{alg:man_sfpca_madmm_full} is initialized at an infeasible pair $(\hat{\bU}, \hat{\bV})$ (unless $\alpha_{\bu} = \alpha_{\bv} = 0$) the first iteration almost always decreases the objective value: after that, however, each sub-problem typically increases the objective value. (If $\lambda_{\bu} = \lambda_{\bv} = 0$, this is guaranteed because the subproblems have a unique global optimum, but we cannot prove monotonicity in general.) Unlike Algorithms \ref{alg:man_sfpca_manpg_full} and \ref{alg:man_sfpca_amanpg_full}, Algorithm \ref{alg:man_sfpca_madmm_full} contains an explicit thresholding step, which we have found improves convergence to exact zeros.

\begin{algorithm}[htb]
\caption{Alternating Maximization Approach for Manifold SFPCA \eqref{eqn:mansfpca} using Manifold ADMM for Subproblem Solutions} \label{alg:man_sfpca_madmm_full}
\begin{enumerate}
  \item Initialize $\hat{\bU}$ and $\hat{\bV}$ as the $k$ leading singular vectors of $\bX$
  \item Repeat Until Convergence:
  \begin{enumerate}
    \item Solve $\bU$-Subproblem using Manifold ADMM \citep{Kovantsky:2016}: \[\hat{\bU} = \argmin_{\bU \in \V^{\bSu}_{n \times k}} -\Tr(\bU^T\bX\hat{\bV}) + \lambda_{\bU} P_{\bU}(\bU) = \argmin_{\bU, \bW_{\bU} \in \V^{\bSu}_{n \times k}} - \Tr(\bU^T\bX\hat{\bV}) + \lambda_{\bU}P_{\bU}(\bW_{\bU}) \text{ subject to } \bU = \bW_{\bU}\]
    \begin{enumerate}
      \item Initialize $\bU^{(k)} = \hat{\bU}$ and restore $\bW^{(k)}_{\bU}, \bZ^{(k)}_{\bU}$ from previous iteration if available else set $\bW^{(k)}_{\bU} = \bU^{(k)}$ and $\bZ^{(k)}_{\bU} = \bzero_{n \times k}$
      \item Repeat Until Convergence:
      \begin{align*}
        \bU^{(k+1)} &= \argmin_{\bU \in \V^{\bSu}_{n \times k}} - \Tr(\bU^T\bX\hat{\bV}) + \frac{\rho}{2}\|\bU - \bW^{(k)}_{\bU} + \bZ^{(k)}_{\bU}\|_F^2 \\
        &= \bS_{\bu}^{-1/2}\bA\bB^T \text{ where } \bA, \bDelta, \bB^T = \SVD(\bSu^{-1/2}\bX\hat{\bV} + \rho\bSu^{1/2}(\bW^{(k)}_{\bU} - \bZ^{(k)}_{\bU})) \\
        \bW^{(k+1)}_{\bU} &= \argmin_{\bW_{\bU} \in \R^{n \times k}} \lambda_{\bU}P_{\bU}(\bW_{\bU}) + \frac{\rho}{2}\|\bU^{(k+1)} - \bW_{\bU} + \bZ^{(k)}_{\bU}\|_F^2 \\
        &= \prox_{\lambda_{\bU}/\rho P_{\bU}(\cdot)}\left(\bU^{(k+1)} + \bZ^{(k)}_{\bU}\right) \\
        \bZ^{(k+1)}_{\bU} &= \bZ^{(k)}_{\bU} + \bU^{(k+1)} - \bW^{(k+1)}_{\bU}
      \end{align*}
    \item Set $\hat{\bU} = \bU^{(k)}$
    \end{enumerate}
    \item Solve $\bV$-Subproblem using Manifold ADMM \citep{Kovantsky:2016}: \[\hat{\bV} = \argmin_{\bV \in \V^{\bSv}_{p \times k}} -\Tr(\hat{\bU}^T\bX\bV) + \lambda_{\bV} P_{\bV}(\bV) = \argmin_{\bV, \bW_{\bV} \in \V^{\bSv}_{p \times k}} - \Tr(\hat{\bU}^T\bX\bV) + \lambda_{\bV} P_{\bV}(\bW_{\bV}) \text{ subject to } \bV = \bW_{\bV}\]
    \begin{enumerate}
      \item Initialize $\bV^{(k)} = \hat{\bV}$ and restore $\bW^{(k)}_{\bV}, \bZ^{(k)}_{\bV}$ from previous iteration if available else set $\bW^{(k)}_{\bV} = \bV^{(k)}$ and $\bZ^{(k)}_{\bV} = \bzero_{p \times k}$
      \item Repeat Until Convergence:
      \begin{align*}
        \bV^{(k+1)} &= \argmin_{\bV \in \V^{\bSv}_{p \times k}} - \Tr(\hat{\bU}^T\bX\bV) + \frac{\rho}{2}\|\bV - \bW^{(k)}_{\bV} + \bZ^{(k)}_{\bV}\|_F^2 \\
        &= \bS_{\bv}^{-1/2}\bA\bB^T \text{ where } \bA, \bDelta, \bB^T = \SVD(\bSv^{-1/2}\bX^T\hat{\bU} + \rho\bSv^{1/2}(\bW^{(k)}_{\bV} - \bZ^{(k)}_{\bV})) \\
        \bW^{(k+1)}_{\bV} &= \argmin_{\bW_{\bV} \in \R^{p \times k}} \lambda_{\bV}P_{\bV}(\bV) + \frac{\rho}{2}\|\bV^{(k+1)} - \bW_{\bV} + \bZ^{(k)}_{\bV}\|_F^2 \\
        &= \prox_{\lambda_{\bV}/\rho P_{\bV}(\cdot)}\left(\bV^{(k+1)} + \bZ^{(k)}_{\bV}\right) \\
        \bZ^{(k+1)}_{\bV} &= \bZ^{(k)}_{\bV} + \bV^{(k+1)} - \bW^{(k+1)}_{\bV}
      \end{align*}
    \end{enumerate}
    \item Set $\hat{\bV} = \bV^{(k)}$
  \end{enumerate}
  \item Return $\hat{\bU}$ and $\hat{\bV}$
\end{enumerate}
\end{algorithm}

\subsection{Additional Note on Identifiability}
As written, the Manifold SFPCA problem \eqref{eqn:mansfpca} suffers from two forms of non-identifiability which may impede convergence, at least in the typical case with $P_{\bU}$ and $P_{\bV}$ elementwise $\ell_1$-penalties \citep{Tibshirani:1996}.  In particular, if the columns of $\bU$ and $\bV$ are simultaneously permuted (\emph{i.e.}, $\bU \to \bU \bS_k$ and $\bV \to \bV \bS_k$ for some permutation matrix $\bS_k$) or if the signs of rows of $\bU$ and $\bV$ are simultaneously flipped, the objective is unchanged. 

To address the first ambiguity (permutation-invariance), the smooth term may be replaced by $\Tr(\bU^T\bX\bV\bD)$ where $\bD$ is a diagonal matrix with elements $(1+\epsilon)^{k-1}, (1+\epsilon)^{k-2}, \dots, 1$ to ensure that the leading PCs are indeed placed first. Alternatively, both problems may be addressed by post-processing the $(\bU, \bV)$ iterates at each step and putting them in a canonical form. A simple canonicalization that we have found works well is to sort the columns of $\bU$ lexographically by absolute value and set the signs so that  the first column of $\bU$ has as many positive elements as possible. 

\section{Additional Experimental Results}
In this section, we give additional details of the simulations performed in Section \ref{sec:sims}.

The timing comparisons of the first simulation (Table \ref{tab:sim1}) are rather sensitive to the specific sub-problem solvers and matrix decomposition subroutines used. For the problem considered, on average MADMM required 30,659 rank-3 SVDs before convergence; ManPG required 439 descent direction solves and 2,259 retractions; and A-ManPG required 366 descent direction solves and 1,952 retractions. Because the cost of an SVD and a retraction (a QR decomposition) are roughly comparable, it is clear that the cost of solving the descent direction subproblem dominates the proximal gradient schemes and that they will be quite sensitive to the solver used. Recently, \citet{Huang:2019} proposed an acceleration scheme for ManPG which could be used here to improve convergence speeds. 

Table \ref{tab:app2} gives an extended version of Table \ref{tab:sim2}, giving additional details on the true and false positive rates of each deflation scheme, as well as measuring subspace recovery performance. For ManSFPCA, the MADMM approach was used to compute the solution, and tuning parameters were fixed as $\lambda_{\bu} = \lambda_{\bv} = 1$ and $\alpha_{\bu} = \alpha_{\bv} = 3$. For the iterative deflation schemes, all four tuning parameters were chosen to maximize the BIC, using the adaptive tuning scheme recommended by \citet{Allen:2019}.

As discussed above, ManSFPCA is able to achieve a greater proportion of variance explained than any of the iterative deflation methods. In terms of subspace recovery, ManSFPCA is far more accurate than the iterative schemes or than an unregularized SVD (indicated by rSS-Error less than 100\%). In terms of variable selection, the BIC scheme used to tune the iterative schemes performs as expected, yielding almost no false positives, at the expense of many false negatives (low TPR). ManSFPCA conversely with fixed tuning parameters has a few more false positives but recovers much more of the true signal than any of the deflation schemes, though this is somewhat sensitive to the choice of tuning parameters. In the orthogonal scenario, the different deflation schemes achieve essentially the same results: in the non-orthogonal scenario, the fuller deflation performed by PD and SD gives a better TPR than Hotelling's deflation. 

\begin{table}[htb]
\centering
\begin{tabular}{ll|cccc|cccc}
\toprule
&&\multicolumn{4}{c|}{\it Scenario 1: $\bU^*$ and $\bV^*$ Orthogonal} & \multicolumn{4}{c}{\it Scenario 2: $\bU^*$ and $\bV^*$ \textbf{Not}  Orthogonal}\\
&& HD & PD & SD & ManSFPCA & HD & PD & SD & ManSFPCA \\
\midrule
\multirow{3}{*}{CPVE} & PC1 & 15.92\% & 21.05\% & \textbf{21.87\%} & \multirow{3}{*}{\bf 37.12\%} & 8.85\% & 19.74\% & \textbf{29.80\%} & \multirow{3}{*}{\bf 50.85\%}\\
& PC2 & 22.21\% & 29.42\% & \textbf{30.59\%} && 13.03\% & 28.30\% & {\bf 39.87\%} \\
& PC3 & 26.80\% & 35.57\% & \textbf{37.09\%} && 16.16\% & 34.22\% & {\bf 46.48\%}\\
\midrule
\multirow{2}{*}{rSS-Error} & $\bU$ & 129.54\% &  129.55\% & 128.35\% & \bf{69.32\%} & 215.73\% & 206.30\% & 205.74\% & \bf{97.77\%}\\
& $\bV$ & 143.01\% & 143.72\% & 141.15\% & \bf{36.98\%} & 211.15\% & 207.77\% & 204.38\% & \bf{78.26\%}\\
\midrule 
\multirow{2}{*}{TPR} & $\bU$ & 54.57\% & 54.29\% & 54.63\% & \textbf{87.72}\% & 11.31\% & 13.11\% & 15.28\% & \textbf{88.91\%}\\
& $\bV$ & 59.20\% & 58.84\% & 59.41\% & \textbf{94.72}\% & 12.89\% & 15.33\% & 18.50\% & \textbf{80.08\%} \\
\midrule 
\multirow{2}{*}{FPR} & $\bU$ & 0.92\% & \textbf{0.91}\% & 0.92\% & 12.28\% & 1.24\% & \textbf{1.05\%} & 1.18\% & 11.09\%\\
& $\bV$ & 0.61\% & \textbf{0.60\%} & 0.60\% & 5.28\% & 2.80\% & \textbf{2.72\%} & 2.79\% & 19.92\% \\
\bottomrule

\end{tabular}
\caption{Extended Version of Table \ref{tab:sim2}, comparing Cumulative Proportion of Variance Explained (CVPE), relative subspace recovery error ($\text{rSS-Error} = \|\hat{\bU}\hat{\bU}^T - \bU^*(\bU^*)^T\| / \|\hat{\bU}_{\text{SVD}}\hat{\bU}_{\text{SVD}}^T - \bU^*(\bU^*)^T\|$), true positive rate (TPR) and false positive rate (FPR) of Manifold SFPCA with Rank-One SFPCA using (normalized) Hotelling, Projection and Schur Complement Deflation.}
\label{tab:app2}
\end{table}

\section{Additional Background}
The literature on regularized PCA variants is vast and we refer the reader to Appendix C of \citet{Allen:2019} for a review.\footnote{Appendix C can be found in the online Supplementary Materials, available at \url{https://arxiv.org/abs/1309.2895}.} As they note, the vast majority of regularized PCA methods use an iterative deflation approach, typically that of \citet{Hotelling:1933}, though the deflations of \citet{Mackey:2008} can be applied to any method which uses the covariance model and the results of Section \ref{sec:deflation} can be used for any which uses the low-rank model. While mainly focusing on the rank-one case, \citet{Journee:2010} show how their modified power algorithm can be extended to simultaneously estimate several orthogonal PCs. \citet{Benidis:2016} give a clever MM-type algorithm for finding orthogonal sparse PCs, based on the iteratively reweighted $\ell_1$-methods first proposed by \citet{Candes:2008} and extended to orthogonality constraints by \citet{Song:2015}. 

In the Bayesian context, the sparse probabilistic PCA model of \citet{Guan:2009} extends the probabilistic PCA model of \citet{Tipping:1999} and allows for sparsity. This method allows for multiple sparse PCs to be estimated simultaneously, though the authors do not discuss specific difficulties associated with orthogonality. While Gibbs sampling from Stiefel-constrained posteriors poses relatively little additional difficulty, it is less obvious how to use more performant MCMC samplers on the Stiefel manifold, especially the popular NUTS variant of Hamiltonian Monte Carlo (HMC) \citep{Neal:2011,Hoffman:2014,Carpenter:2017,Betancourt:2017a}. \citet{Girolami:2011} and \citet{Byrne:2013} propose geodesic variants of HMC, which extend the strong geometric foundations of (Euclidean) HMC \citep{Betancourt:2017b} to arbitrary smooth manifolds, but they are not easy to apply in practice and robust software implementations have yet to be developed. Working around this, several authors have proposed methods to reparameterize the Stiefel manifold into an approximately Euclidean coordinate system and apply standard HMC, though the practicality of these methods is limited by the incompatible structures of the Stiefel manifold and Euclidean space \citep{Pourzanjani:2017,Janch:2018}. More recently, \citet{Janch:2019} proposed a promising data augmentation scheme based on the polar decomposition which avoids these topological inconsistencies, at the cost of a slightly larger parameter space.

The use of manifold optimization techniques for estimating multiple regularized PCs simultaneously is a recent development, driven by recent developments in non-smooth manifold optimization, especially the proximal gradient scheme of \citet{Chen:2018,Chen:2019} and the Manifold-constrained ADMM of \citet{Kovantsky:2016}, as well as references therein. (\citet{Lai:2014} give an interesting splitting method for Stiefel-constrained problems based on Bregman iteration techniques, but to the best of our knowledge, it has not yet been applied to sparse PCA formulations.) Theory for these methods is still under rapid development and several key results remain unproven: in particular, convergence of the Manifold ADMM \citep{Kovantsky:2016} has not been established, though the analysis of \citet{Wang:2019} comes somewhat close. 

On the contrary, smooth manifold optimization techniques are well-established and typically motivated by problems in physics and engineering, where conservation laws are expressed as manifold constraints. These techniques were (re-)popularized by the influential book of \citet{Absil:2007}. \citet{Wen:2013} give a particularly nice algorithm which maintains feasibility by using a clever application of the Cayley transform. \citet{Ritchie:2019}  used smooth manifold optimization to solve a supervised PCA problem. Interestingly, their approach is Grassmanian- rather than Stiefel-constrained because, in a non-sparse context, the specific coordinate system used is irrelevant. This allows them to use the Grassmanian-based techniques of \citet{Edelman:1998} and of \citet{Boumal:2018}.

\section{Additional References}
\printbibliography[heading=none]

\end{refsection}
\end{document}